%% file: main.tex
\documentclass[11pt, a4paper, logo, twocolumn]{berkeley}

\usepackage{moreverb,url}
\usepackage{array}    
\usepackage{tabularx} 
\newcolumntype{Y}{>{\centering\arraybackslash}X}
\usepackage{afterpage}
\usepackage{amssymb}
\usepackage{pifont}
\usepackage{array}
\usepackage{subcaption}
\newcolumntype{C}[1]{>{\centering\arraybackslash}b{#1}}

\newcommand\BibTeX{{\rmfamily B\kern-.05em \textsc{i\kern-.025em b}\kern-.08em
T\kern-.1667em\lower.7ex\hbox{E}\kern-.125emX}}

\usepackage[all]{hypcap}
\usepackage{xspace}
\usepackage[capitalize,noabbrev]{cleveref}
\usepackage{subcaption}
\usepackage{wrapfig}
\usepackage{lipsum}

\makeatletter
\def\mathcolor#1#{\@mathcolor{#1}}
\def\@mathcolor#1#2#3{%
  \protect\leavevmode
  \begingroup
    \color#1{#2}#3%
  \endgroup
}
\makeatother
\renewcommand{\today}{July 2024}

\title{FMB: a Functional Manipulation Benchmark for
Generalizable Robotic Learning }
\runningtitle{FMB: a Functional Manipulation Benchmark for
Generalizable Robotic Learning }

\author[1*]{Jianlan Luo}
\author[1*]{Charles Xu}
\author[1]{Fangchen Liu}
\author[1]{Liam Tan}
\author[1]{Zipeng Lin}
\author[1]{Jeffrey Wu}
\author[1]{Pieter Abbeel}
\author[1]{Sergey Levine}

\affil[*]{Equal Contribution}
\affil[1]{Department of Electric Engineering and Computer Sciences, University of California, Berkeley, USA}

\correspondingauthor{Jianlan Luo, Berkeley AI Research(BAIR), University of California, Berkeley, 2121 Berkeley Way West, Berkeley, CA, 94704, USA \href{mailto:jianlanluo@eecs.berkeley.edu}{jianlanluo@eecs.berkeley.edu}}

\begin{abstract}
In this paper, we propose a real-world benchmark for studying robotic learning in the context of functional manipulation: a robot needs to accomplish complex long-horizon behaviors by composing individual manipulation skills in functionally relevant ways.
The core design principles of our Functional Manipulation Benchmark (FMB) emphasize a harmonious balance between complexity and accessibility. Tasks are deliberately scoped to be narrow, ensuring that models and datasets of manageable scale can be utilized effectively to track progress. Simultaneously, they are diverse enough to pose a significant generalization challenge. Furthermore, the benchmark is designed to be easily replicable, encompassing all essential hardware and software components.
To achieve this goal, FMB consists of a variety of 3D-printed objects designed for easy and accurate replication by other researchers.
The objects are procedurally generated, providing a principled framework to study generalization in a controlled fashion.
We focus on fundamental manipulation skills, including grasping, repositioning, and a range of assembly behaviors. The FMB can be used to evaluate methods for acquiring individual skills, as well as methods for effectively combining and ordering such skills in order to solve complex, multi-stage manipulation tasks.
We also offer an imitation learning framework that includes a suite of policies trained to solve the proposed tasks.
This enables researchers to utilize our tasks as a versatile toolkit for examining various parts of the pipeline. For example, researchers could propose a better design for a grasping controller and evaluate it in combination with our baseline reorientation and assembly policies as part of a pipeline for solving multi-stage tasks.
Our dataset, object CAD files, code, and evaluation videos can be found on our project website: \url{https://functional-manipulation-benchmark.github.io}.
\end{abstract}

\begin{document}

\maketitle
\input{sections/intro.tex}

\input{sections/related_work}
\input{sections/benchmark.tex}
\input{sections/usage.tex}
\input{sections/experiment.tex}
\input{sections/discussion}

\section*{Acknowledgments}This research was partly supported by the National Science Foundation through IIS-2150826, ONR through N00014-22-1-2773 and N00014-20-1-2383, Intrinsic, and the AI Institute. We also thank the computing resource support from the Berkeley Research
Computing (BRC) program, the NSF Cloudbank program, and the Google TPU Research Cloud program.

\bibliography{ref}

\end{document}

%% file: sections/intro.tex
\section{Introduction}\label{sec:intro}
\begin{figure*}[t]
    \centering
    \includegraphics[width=\textwidth]{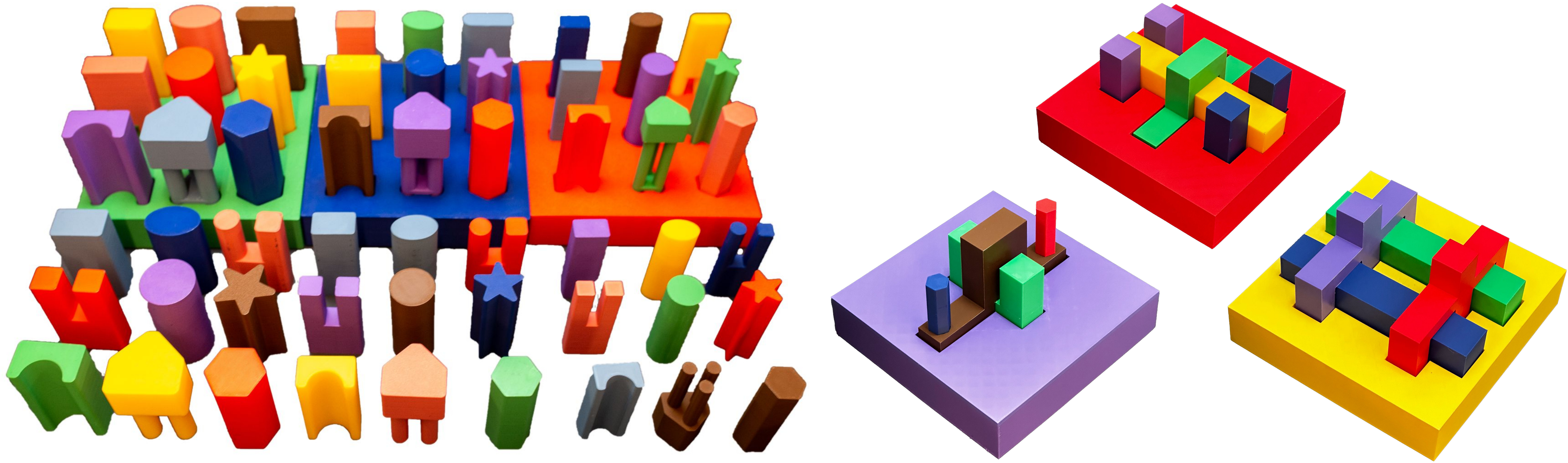}
    \caption{\textbf{Left:} The 3D-printed parts for single-object manipulation tasks. \textbf{Right:} Three instantiations of the complex assembly task. These tasks require similar functional manipulation behaviors as the simpler set of tasks but with multiple interlocking objects and a more complex higher-level structure that requires assembling the parts in the right order.}
    \label{fig:teaser_fig}
\end{figure*}

Manipulation is one of the foundational problems in robotics research, but enabling robots to perform dexterous manipulation skills that reflect the capabilities of humans is still out of reach. In fact, even matching the performance of human \emph{teleoperation} remains a major challenge, particularly in environments that require generalization and are not constrained to a specific fixed set of well-characterized objects.
As \citet{cui_sci_review} point out, two primary difficulties in robotic manipulation lie in intelligently handling complex contact dynamics and the variability in the environment and objects. 
Robotic learning techniques hold the potential to address these challenges. However, making effective and measurable progress will require a comprehensive and accessible framework to offer essential components: sufficiently challenging tasks of practical relevance, reasonable amounts of high-quality data, an easy-to-reproduce setup, a collection of relevant methods providing baseline results, and thorough analysis of the experimental findings on the proposed tasks.

\begin{figure*}[ht!]
    \centering
    \includegraphics[width=\textwidth]{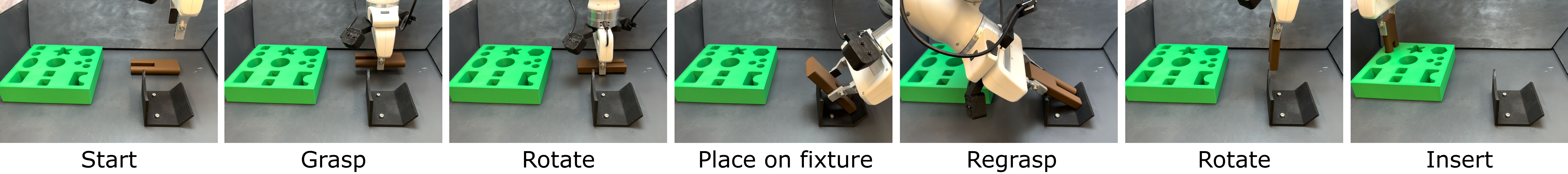}
    \vspace{-2em}
    \caption{An illustration of the steps for completing a Single-Object Manipulation Task, which requires grasping the part, reorienting it (potentially using an environment fixture), and then inserting it into the appropriate slot.}
    \label{fig:single-object-filmstrip}
\end{figure*}

\begin{figure*}[ht!]
    \centering
    \includegraphics[width=\textwidth]{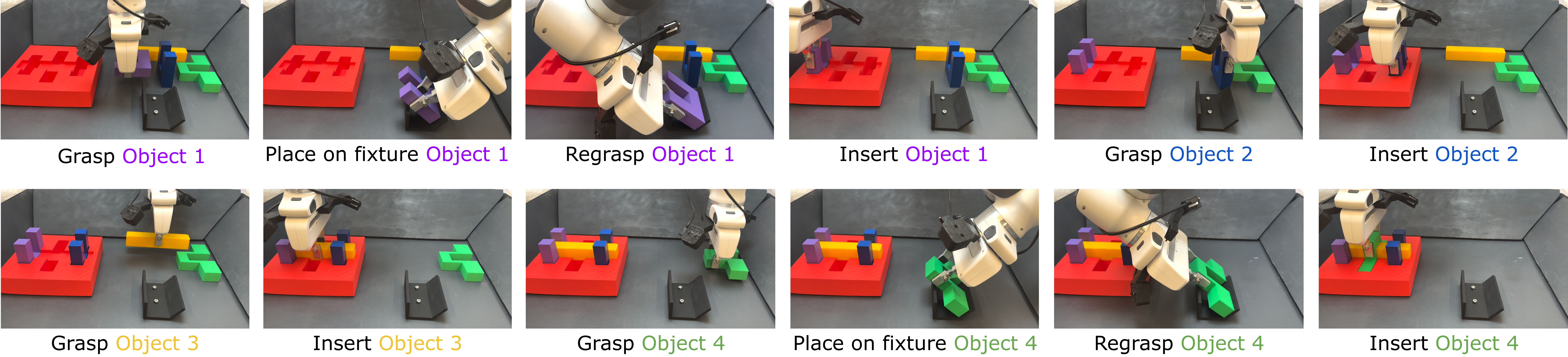}
    \vspace{-2em}
    \caption{An illustration of the steps for solving a Multi-Object Manipulation Task, which requires performing the same skills as the Single-Object Task repeatedly for each component in the interlocking assembly.}
    \label{fig:multi-object-filmstrip}
\end{figure*}
While significant recent research in robotic learning has made progress on various aspects of manipulation problems~\citep{levine2016end, kalashnikov2018qtopt, rt12022arxiv, zeng2020transporter, xu2022dextairity, doi:10.1177/027836499101000105, gu2016deep, schaal2007, schaal2011, abbeel2008nips, billard2018, mahler2017dexnet}, 
much of the emphasis in recent works have either been on broad generalization with relatively simple skills, which often do not capture many physical challenges of manipulation (e.g. imprecise pick-and-place tasks)~\citep{pinto2015supersizing, levine2016learning, ebert2021bridge}, or performing narrow tasks with physically more complex skills without extensive generalization~\citep{openai2019solving, nist, 8794074,hu2023reboot}. This is not unreasonable: it is very difficult to simultaneously make progress on broad generalization (which often requires huge datasets) and tackle the full physical complexity of dexterous manipulation. So how can we take a step toward facilitating robotic learning research that emphasizes both generalization and physically intricate skills while still keeping the problem constrained enough so as to enable meaningful progress?

In this paper, we propose such a real-world benchmark, which we call the functional manipulation benchmark (FMB).
FMB aims to cover important dimensions of physical complexity and object generalization while still providing a degree of accessibility by carefully restricting the scope to a domain where we can make progress with reasonably sized datasets and models. 
We approach the design of this benchmark by defining functional manipulation as the problem of manipulating objects in ways functionally relevant to a sequence of manipulation behaviors, such as picking up an object with an appropriate pose, repositioning it if necessary, and then using it for physical interactions. Two such examples can be seen in Fig.~\ref{fig:single-object-filmstrip} and Fig.~\ref{fig:multi-object-filmstrip}.
While this definition is more restrictive, we believe it captures a broad range of practical manipulation tasks and includes both the challenges of contact dynamics and object generalization.

The specific tasks we instantiate to capture functional manipulation are themed around assembly problems, including pick-and-place tasks and more complex long-horizon multi-stage multi-part assemblies. 
These tasks, illustrated in Fig.~\ref{fig:teaser_fig}, require picking up the individual pieces, reorienting them (potentially using environment fixtures and regrasping), and then slotting them into their corresponding location. 
Each phase requires addressing the challenge of complex contact dynamics, skill sequencing strategies, as well as object generalization. 
The objects vary in shape, size and color between training and testing phases, and their locations are randomized. The grasping phase requires selecting a grasp that is suitable for reorienting or inserting the object, the reorientation phase requires positioning the object so that its pose can be adjusted in the desired way, and the assembly phase requires compliant insertion and proper accounting for the contact forces on the object. Each phase requires handling different objects (including held-out objects) and different poses. 
The overall sequencing strategy needs to serve as the mechanism of composing such skills appropriately, as well as recovering from failed execution. For example, for the task presented in Fig.~\ref{fig:single-object-filmstrip}, the robot may need to retry grasping on failed ones multiple times until it firmly holds the object before advancing to the next stage. In tasks illustrated in Fig.~\ref{fig:multi-object-filmstrip}, the robot must further reason the right sequence of manipulation as these objects are assembled in an interlocking fashion. 

To ensure the reproducibility and portability of such tasks, we designed 66 3D-printed objects with diverse shapes and sizes that can be easily replicated by other researchers.
Accompanying these objects, we collected a dataset of 22,500 human demonstrations of grasping, repositioning, and assembly skills.
Our dataset contains a variety of sensory modalities, as presented in Fig.~\ref{fig:setup}: we record RGB and depth images from multiple cameras, relevant robot kinematics information, as well as force/torque measurement at the robot's end-effector frame. 
We also trained a set of imitation learning policies to perform either individual stages or the entire assembly tasks. These policies are also provided as pre-trained model checkpoints so that they can be reused by others as component parts of larger systems or as scaffolds for studying improvements to individual stages.
FMB is modular so that other researchers can repurpose it for a variety of methods that they may wish to develop and can focus on any stage or aspect of the task. For example, some researchers might choose to focus on better functional grasping or assembly methods, while the other stages are handled by our baseline system. Some researchers might focus on skill sequencing, utilizing trained skills from our system for the individual steps. Others might also focus on developing an end-to-end method for the entire multi-stage task, fully utilizing the provided training data. 
With the accessible and extensive framework that FMB provides, our hope is that it can serve as a ``toolkit'' to facilitate the entry of researchers into the field of robot learning with ease.

%% file: sections/related_work.tex
\section{Related Work}\label{sec:related_work}

Considerable recent progress in robotic manipulation has studied generalization, though often in the context of simpler tasks such as grasping~\citep{dasari2020robonet, levine2016learning, yang2019replab}, pushing~\citep{dasari2020robonet, finn2016pushing}, and imprecise repositioning~\citep{dasari2020robonet, lee2021rgbstacking}. A number of other works have studied tasks that are dynamic~\citep{seita2019deep}, precise (e.g., insertion)~\citep{zakka2020form2fit}, contact-rich~\citep{benchmarkforce2016}, or otherwise physically challenging~\citep{openai2019solving, nistboard}. Fewer works have studied these factors in combination~\citep{heo2023furniturebench}. We believe many of the central challenges in robotic manipulation lie at the confluence of these two challenges: tasks that require handling contact dynamics, not by memorizing the particular pattern needed for a single narrow task, but by learning general behaviors for handling object interaction that can generalize to new objects. Our aim is to propose a benchmark that can study this combination of challenges while keeping the scope narrow enough that it remains accessible to many researchers.

Our functional manipulation tasks combine aspects of grasping, repositioning, and assembly. A number of works have studied functional grasping~\citep{levine2016learning, 4600696, 769, liu2020cage, zhao2020robotic}, and insertion \citep{mahler2017dexnet} separately. Our goal is not to attain the best possible performance in narrow settings for any of these stages (e.g., ultra-high-precision industrial insertion e.g., NIST board challenge~\citep{nist}) but to use these tasks as a lens through which to gauge general manipulation capabilities learned via general-purpose robotic learning methods.

A number of prior works have proposed datasets for robotic learning, including datasets consisting of demonstrations~\citep{ebert2021bridge, walke2023bridgedata, fang2023rh20t} and autonomously collected data~\citep{levine2016learning, pinto2015supersizing}, as well as annotated datasets of grasp points~\citep{fang2020graspnet}, object geometries~\citep{tyree20226dof, p2023fewsol}, simulated environments~\citep{james2019simtoreal}, and multimodal inputs~\citep{fang2023rh20t}. However, there has been comparatively little work on standard and accessible object sets that are combined with multi-stage tasks for studying generalization. The YCB object set comes with a number of evaluation protocols~\citep{Calli_2015}, but these protocols generally focus on object repositioning tasks that do not evaluate the complex contacts challenges that we discuss in the previous section. A number of existing demonstration datasets cover many different behaviors~\citep{ebert2021bridge, mandlekar2019scaling,walke2023bridgedata,dasari2020robonet,bharadhwaj2023roboagent}, but also focus more on behaviors that emphasize basic pick-and-place skills rather than precise or contact-rich manipulation. Some works have focused on insertion skills in particular (e.g., connector insertion)~\citep{demagistris2018experimental,luo2019reinforcement,Luo-RSS-21,zhao2022insertion, tangpeghole,Yun2008, Bruyninckx}. While FMB is related, we aim specifically to cover a range of skills, including grasping and repositioning, that we believe cover a basis of basic manipulation capabilities. We also emphasize generalization as a primary challenge for FMB.

We use 3D-printed objects to facilitate reproducibility. Other prior works have also proposed standard meshes and 3D printed parts for benchmarking and reproducibility~\citep{Calli_2015}, typically focusing on object grasping. These efforts are related, but our aim is to provide parts that are specifically well suited for evaluating all of the stages: grasping, reorientation, and assembly, rather than only grasping. 

%% file: sections/benchmark.tex
\section{Functional Manipulation Benchmark}
\label{sec:fmb}

In this section, we introduce the basic principles behind FMB and the protocols to evaluate different methods on this benchmark. FMB tasks can broadly fall into two categories: single-object multi-stage manipulation tasks and multi-object multi-stage manipulation tasks. They both require acquiring individual manipulation skills such as grasping, repositioning, and insertion, as well as composing these individual skills to complete the full task as depicted in Fig.~\ref{fig:single-object-filmstrip} and Fig.~\ref{fig:multi-object-filmstrip}. These two categories bear similar design principles but differ in the additional complexity of the second category, which involves selecting the appropriate object for manipulation.
We are primarily concerned with studying the generalization of each individual functional manipulation skill as well as evaluating the performance of different methods on the full assembly task.
Therefore, we collect a diverse dataset of robotic behaviors with different objects, viewpoints, and robot initial poses. We also provide novel objects to evaluate the generalization capability of individual skills. Thus, we test the generalization of learned manipulation skills in terms of object location and physical attributes.

\subsection{Object Set}

The objects in FMB are 3D-printable, and the CAD files are available on our website. In total, we have 66 objects as in Fig.~\ref{fig:teaser_fig}, 54 of them belong to single object manipulation tasks; the remaining compose the multi-object manipulation tasks. 
Out of these 54 objects, we designed nine different basic shapes and six different sizes for each shape; each object is assigned one of eight colors specified on our website. These objects are paired with three boards with matching openings as in the left of Fig.~\ref{fig:teaser_fig}.
We additionally designed three more complex boards to facilitate multi-stage assembly tasks, shown in the right of Fig.~\ref{fig:teaser_fig}; objects there are generated procedurally so that they can only be fit together in specific orders.
The tolerance for mating all objects is between 1mm and 2mm, which is practical for commercial 3D printers available on the market. Additionally, we created 5 test objects used to evaluate the generalization capabilities. These vary in shape, size, and color from the training objects.

\begin{figure*}[ht!]
    \centering
    \includegraphics[width=\textwidth]{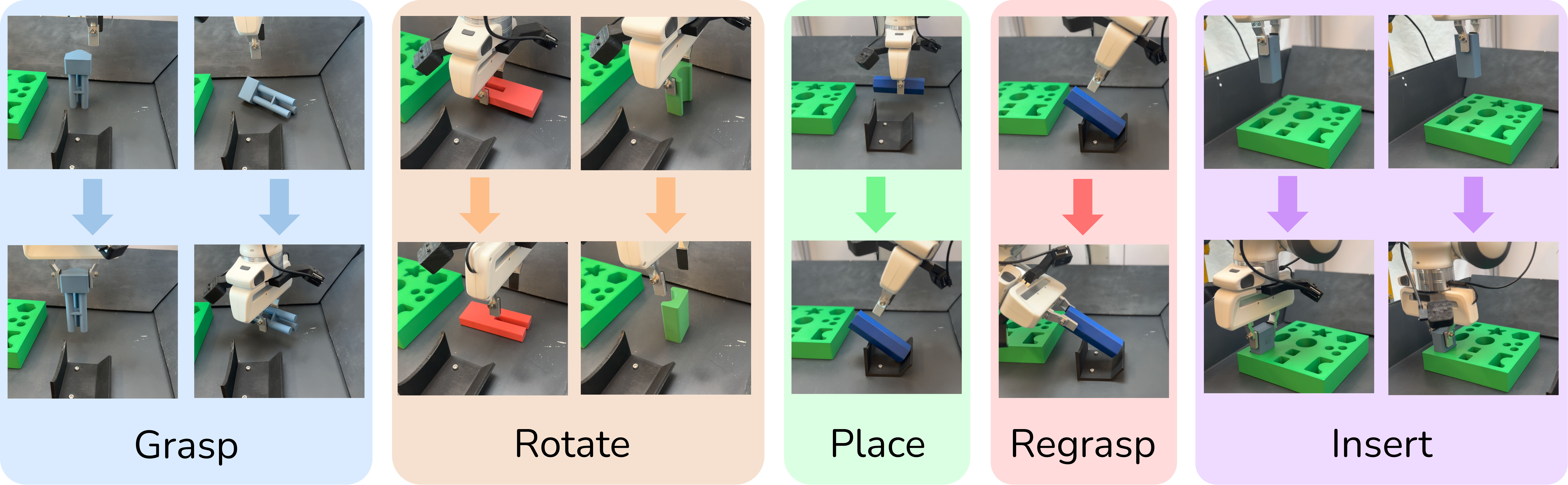}
    \caption{Illustration of individual skills in the Single-Object Task. Note that the grasp and rotate skills have to manipulate the object in both the vertical and horizontal orientation. For isometric shapes like the rectangle, the insert skill needs to decide whether to rotate the object to line up with the hole.}
    \vspace{-1.2em}
    \label{fig:primitives}
\end{figure*}

\subsection{Individual Manipulation Skills}

In this section, we describe the ``primitive'' manipulation skills included in FMB for evaluation as well as our data collection system. For each type of skill, we provide demonstration trajectories collected with a Franka robot (see Fig.~\ref{fig:setup}) and an evaluation protocol as in Section~\ref{sec:protocol}. This modular design of our benchmark facilitates extension to add new tasks with the provided objects, and the tasks we describe here are suitable both for evaluating generalization and for testing a range of manipulation capabilities.

\paragraph{Grasping.} The grasping task in our benchmark is a \emph{functional} grasping task, in the sense that the robot must grasp the object in a way that facilitates downstream manipulation rather than simply picking the object in any pose.
We illustrate this task in Fig.~\ref{fig:primitives}. 
For example, if we are going to perform insertion after grasping an object, a top-down grasp is reasonable if the object is placed in a vertical pose, as shown on the right side of Fig.~\ref{fig:primitives}. However, a horizontal grasp is much more desirable if the object is positioned as in the second row of the left side of Fig.~\ref{fig:primitives}; because it can be impossible to find a collision-free path to grasp vertically on the top of the blue object or easily violating the robot's kinematics constraints to perform downstream manipulation even if such grasps can be found.
In such scenarios, the robot needs to perform additional repositioning steps to adjust the feasible grasp pose. The robot must learn grasping skills that deploy the appropriate grasp conditioning on the object's current configuration and also generalize across different object shapes, colors, and sizes. Our demonstration dataset for the grasping task consists of 50 trajectories per object, with varying object rest poses in the randomization zone, for a total of 2700 trajectories performing functional grasping over the 54-object set; additionally, we collect 1800 grasping trajectories for objects in the three multi-object assembly tasks, so each object gets 150 demonstrations in a randomized setting of placements among other objects.

\paragraph{Repositioning.} A repositioning step is sometimes necessary to adjust the grasping pose so that the object is held in a way that is suitable for downstream assembly, as mentioned in the last paragraph. 
For objects with asymmetric geometries, a rotation operation is usually desirable for the downstream insertion task. For example, the objects in the second column of Fig.~\ref{fig:primitives} need to be rotated 180 degrees so that they can slot into the matching holes in the board more easily. 
On the other side, manipulating and reorienting objects by leveraging environment affordances (e.g., tilting the object in the gripper by levering it against a table or wall) may often be necessary for fluent and complex manipulation, and this reorientation task exercises this capability. 
We provide a simple fixture that can serve as an environment affordance to rest the object at an angle, as shown in Fig.~\ref{fig:primitives}. To reorient the objects into the right pose, the robot may need to use this fixture, resting the object on it and then regrasping it in a more appropriate pose for reorientation. We collected 4500 demonstrations for placing and regrasping, which can be used to learn strategies for using environmental affordances for regrasping and reorientation. 
Since objects land in the fixture in a relatively deterministic fashion, we partially script our demonstration collection process while maintaining a certain degree of randomness for the purpose of data diversity. We detail such process and code of implementing it on our website.

\paragraph{Insertion.} Our assembly tasks require inserting objects with diverse shapes into their matching slots, which requires performing fine-grained precise manipulation. An illustrative example is shown in Fig.~\ref{fig:primitives}. Here, having completed the preceding steps, the robot is holding an object and needs to insert it into the matching slot on the board. 
For the single-object task, we collected 125 human demonstrations that include various robot initial poses and board positions, for a total of 6750 demonstrations performing the assembly task from various initial conditions. 
Note that in the single-object task, the board's pose is randomized within a 35 x 35 cm region and rotated up to 15º in each episode, requiring a reactive strategy that localizes the board and the appropriate matching slot, and guides the object into the correct location. Similarly, 150 human demonstrations were collected per object in the multi-object assembly tasks, resulting in a total of 1800 trajectories.

\subsection{Single-Object Multi-Stage Manipulation Tasks}
Aside from performing individual steps mentioned above, such as grasping, reorientation, and assembly, our benchmark and demonstrations can be used to learn the entire long-horizon sequence, composing these steps to insert a free object into the assembly board; one such example is shown in Fig.~\ref{fig:single-object-filmstrip}.
The difficulty of this task mainly comes from the compounding errors accumulated over each individual step which gets even more magnified when switching between tasks. For instance, after completing the grasping and repositioning stages, an object might be held in a pose different from the ones in the human demonstration data used for insertion.

\begin{figure}[t!]
    \centering
    \includegraphics[width=1\linewidth]{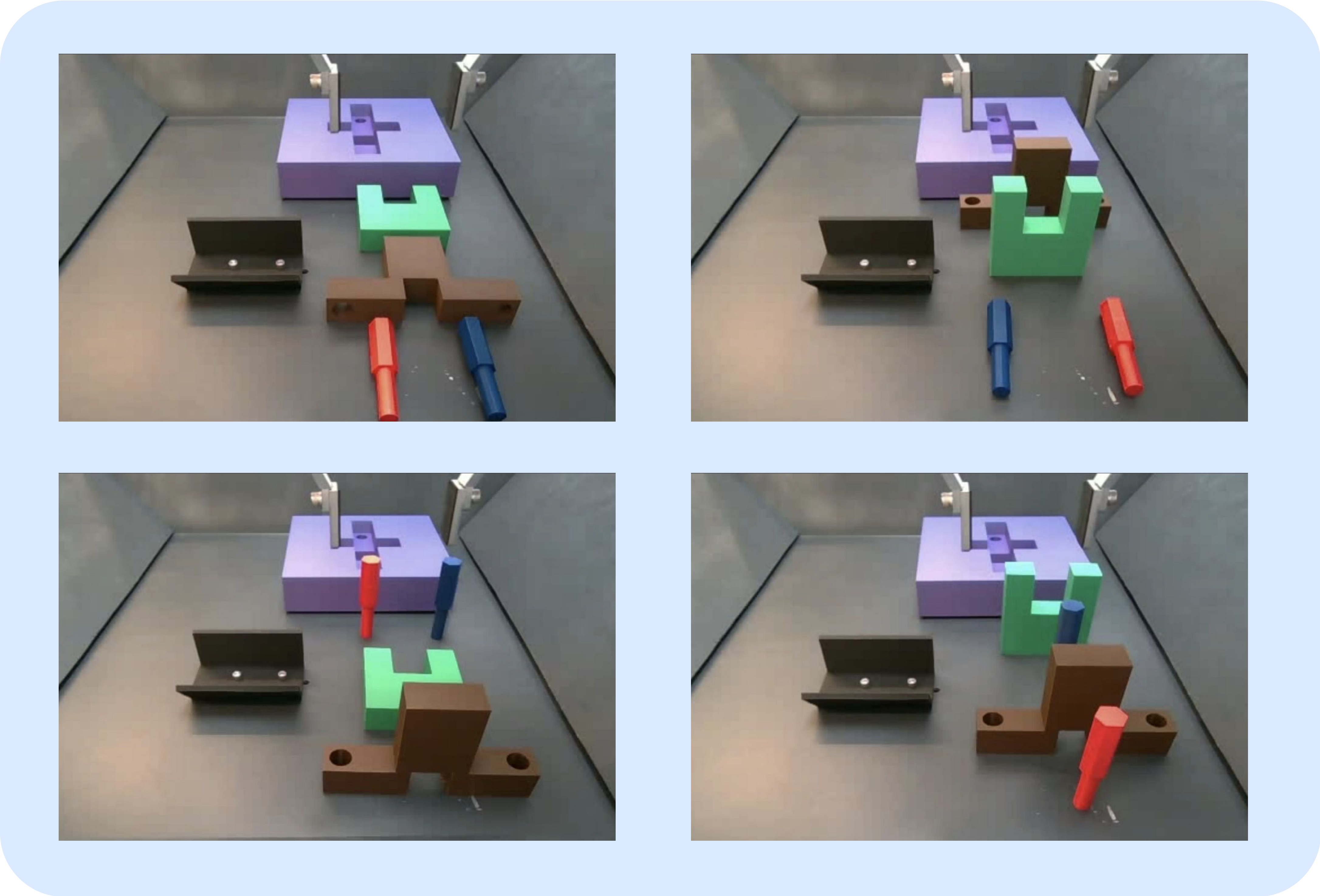}
    \caption{Example of different initial configurations for one Multi-Object Task assembly board. We randomize both the orientation and position relative to other objects at the start of each assembly demonstration episode.}
    \label{fig:board_grasp_init}
\end{figure}

\begin{figure}[h]
    \centering
    \includegraphics[width=0.9\linewidth]{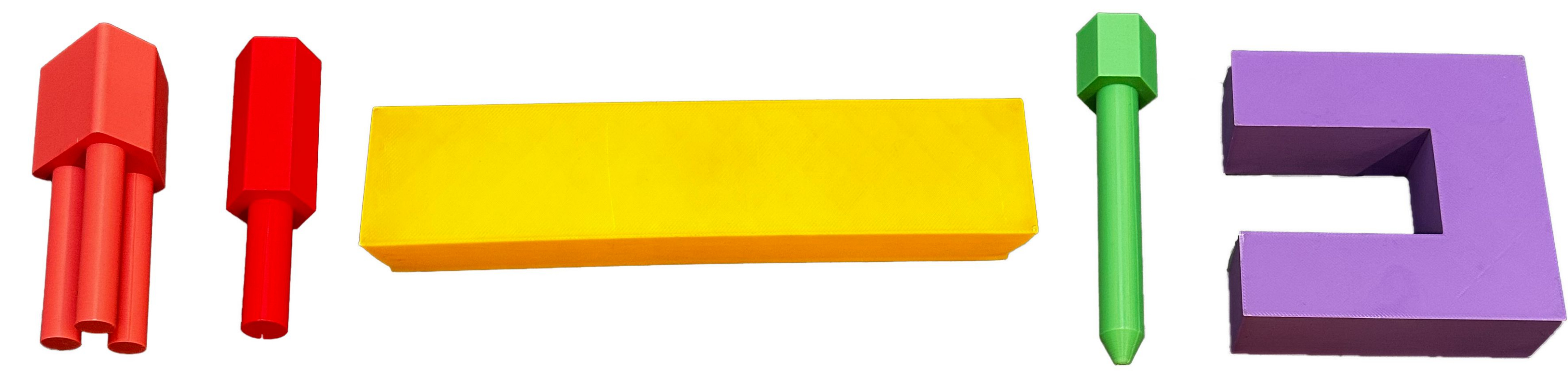}
    \caption{Unseen test objects used for evaluating generalization to new combinations of shapes, sizes, and colors. }
    \label{fig:novel}
\end{figure}

\subsection{Multi-Object Multi-Stage Manipulation Tasks} 
\begin{figure}[t!]
    \centering
    \includegraphics[width=1\linewidth]{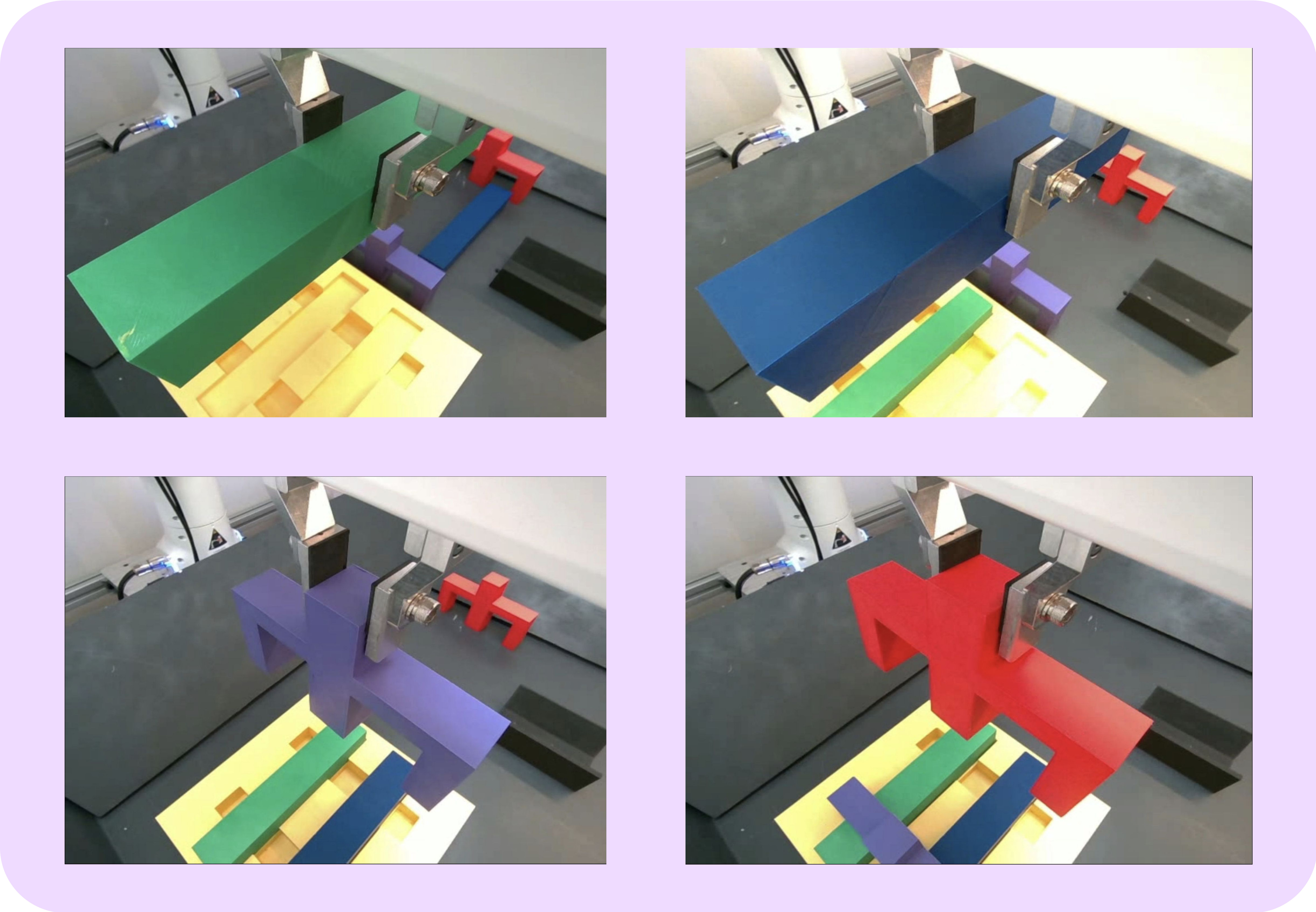}
    \caption{Various initial distributions of the insert skill for the Multi-Object Task. In each instance, there are different numbers of objects already inserted into the board. }
    \label{fig:board_insert_init}
\vspace{-0.2cm}
\end{figure}

We also present three sets of more challenging objects for assembly, as presented in the right of Fig.~\ref{fig:teaser_fig}. 
These tasks are more challenging than the single-object tasks since the pieces fit in an interlocking fashion, so there is much more variability in which object to perform manipulation skills on. 
For the grasping stage, as pictured in Fig.~\ref{fig:board_grasp_init}, the robot needs to grasp a desired object among several others with the added complexity of randomized object placements for each attempt.
For the insertion stage, as illustrated in Fig.~\ref{fig:board_insert_init}, the robot needs to insert objects while coming into contact with other objects already present on the assembly board. This situation introduces more complexity in contact dynamics, necessitating a higher level of precision in manipulation.
Another major challenge with these tasks is that the interlocking pieces need to be put together in a specific order. While it may not be too hard to perform individual steps alone, the difficulty increases rapidly when a policy needs to simultaneously reason the manipulation sequence as well as accounting for compounding manipulation errors introduced by individual steps.

\begin{figure*}[t]
    \centering
    \includegraphics[width=\textwidth]{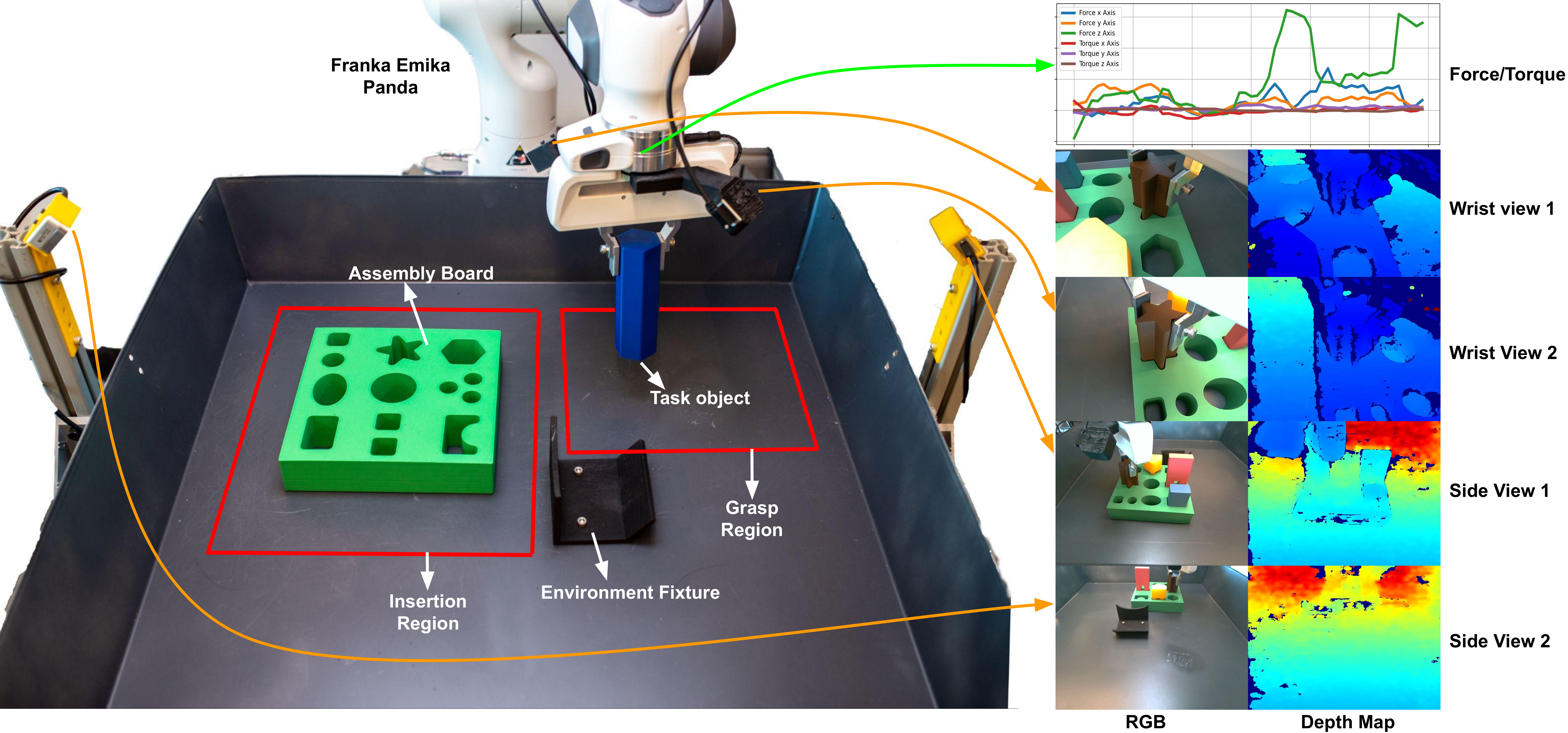}
    \caption{Illustration of the robot setup, with a standard Franka arm equipped with four cameras (two on the wrist and two attached to the environment), each with RGB and depth channels, positioned in front of a workspace containing an object, reorientation fixture, and assembly board. The board is placed into a random pose within the randomization region, and the object is located in a randomized pose on the table, from where it must be picked up, reoriented, and inserted.}
    \label{fig:setup}
 \vspace{-0.2cm}
\end{figure*}

\begin{table*}[t]
    \centering
    \begin{tabular}{c c c c c c c c}
    \hline
        & Grasp & Place\_on\_fixture & Regrasp & Rotate & Move\_to\_board &Insert & \textbf{Total} \\
        Single-Object Task &  2,700 & 1,350 & 1,350 & 500 & 
 2,700 & 6,750 & \textbf{15,350}\\
        Multi-Object Task & 1,800 & 900 & 900 & 0 & 1,800 & 1,800 & \textbf{7,200}\\
    \hline
        \textbf{Subtotal} & 4,500 & 2,250 & 2,250 & 500& 4,500 & 8,550 & \textbf{22,550} \\
    \hline
    \end{tabular}
    \caption{Number of demonstration trajectories in our dataset separated by primitive and task. Each trajectory is approximately 5 seconds in length, for a total of 22,550 trajectories.}
    \label{tab:num_traj}
\vspace{-0.2cm}
\end{table*}

\subsection{Robotic System and Data Collection}
We now describe the robotic system and details of the 22,550 demonstration trajectories that we collected and released as part of the benchmark. The dataset composition can be seen in Table~\ref{tab:num_traj}.

\paragraph{Robotic system overview}
Our system can be seen in Fig.~\ref{fig:setup}. We use a Franka Panda robot to collect our dataset since it is widely adopted for research and offers a torque control interface which is very desirable in contact-rich manipulation tasks. 
To tele-operate the robot, we use a SpaceMouse to command 6 DoF end-effector twist at 10 Hz, which is then tracked by a low-level impedance controller running at 1K Hz. The software for operating the robot, as well as the low-level controller, is also included in our open-sourced release.
In total, we have four Intel RealSense D405 cameras, two of which are mounted on the robot end-effector, and the rest are placed on each side of the bin to provide a complementary view of objects in the bin. To ensure the image observations are free of background distractions, we put white curtains around the side of the workspace.
We simultaneously capture RGB and depth images from these cameras, and we also provide calibrated camera intrinsics. This calibration allows for the conversion of depth images into point clouds when necessary.
We also log the end-effector force/torque information provided by the Franka Panda robot. We did not use an additional force/torque sensor as it simplifies the standardization process by utilizing the robot's inherent sensing capabilities\footnote{The Franka Panda robots utilize a computational model to estimate the force and torque at the end-effector, rather than direct sensory measurements. According to the user manual, the force resolution is 0.05N, and the torque resolution is 0.02Nm; we found the quality of the readings is sufficient for FMB.}. 
Our robotic system setup is simple and modular; one can reproduce our exact setup by following the procedure on our website \url{https://functional-manipulation-benchmark.github.io/files/index.html}.

\paragraph{Single-object task dataset.}
Our dataset comprises 2700 demonstrations of the complete single-object task, encompassing every aspect from grasping and reorientation to object insertion. Each stage within these complete trajectories is automatically labeled, enabling the segmentation of trajectories into individual skills by querying the corresponding labels.
We also collected an additional 4050 demonstrations of the insertion stage alone since it's a much harder task, thus requiring more data.
Each end-to-end demonstration trajectory ranges from 20 to 40 seconds in length. One can directly learn a ``flat" policy on these long trajectories or break them into ``primitive" trajectory sequences using the labels mentioned before.
In our dataset, these primitives include \texttt{grasp}, \texttt{place on fixture}, \texttt{regrasp}, \texttt{rotate}, \texttt{move to board}, and \texttt{insertion}. 
After segmenting by primitives, we end up with a total of 15,350 demonstrations, with an average length of about 5 seconds.
As shown in Fig. \ref{fig:setup}, the pose of the task object for the grasping task is randomized around a 20cm$\times$20cm rectangular area in the bin. For the insertion task, the board is randomized inside a 35cm$\times$35cm area. A drawing of such a protocol can be found on our website. We also include distractors (i.e., objects not needed for a task) when performing the insertion task. One-fifth of the insertion demonstrations were carried out when there were distractors present.

\paragraph{Multi-object task dataset.}
In addition to the single-object manipulation task dataset, we also collected 150 end-to-end demonstrations of solving each of the three multi-object assemblies. Each trajectory contains steps to grasp, reorient, and insert the four components of the assembly sequentially and can exceed 100 seconds in length. We again break them down into separate primitives like \texttt{grasp}, \texttt{place on fixture}, \texttt{regrasp}, \texttt{move to board}, and \texttt{insert} for each manipulation object. After segmentation, this part of the dataset contains 7,200 trajectories with lengths of about 5 seconds.

For the multi-object manipulation task, all four assembly objects are randomly placed in the 20cm$\times$30cm area, requiring the learned system to determine the desired piece to pick up. Unlike the single-object task, the assembly board is fixed to the table. A drawing of such a protocol can be found on our website.

\subsection{Using the Benchmark}
To use the FMB benchmark, users would first need to reproduce the setup. This includes purchasing relevant equipment, such as the bin and cameras, as well as printing the FMB objects and tools with specified materials and colors. The detailed instructions can be found on our website \url{https://functional-manipulation-benchmark.github.io/usage/index.html}. Our dataset was collected using a Franka panda robot; however, users could still use relevant components within the FMB framework to collect their own data if they choose to use a different robot.


\subsection{Evaluation protocol}
\label{sec:protocol}
In order to evaluate the performance of different methods, we designed a set of detailed evaluation protocols for each task of FMB.
In these protocols, we specify a set of object initial poses within the randomization region to test the proposed methods' generalization capability while ensuring consistency across different experiments and labs.

\paragraph{Single-object tasks.} For grasping and repositioning tasks, one can hold out a specific object in the training set, train a policy without seeing any data associated with that object, and then test on the held-out object. Additionally, we also provide novel objects that are not contained in the dataset for which researchers can directly evaluate the trained policies, such as the five objects shown in Fig.~\ref{fig:novel}.
Furthermore, we define a set of specific starting poses for both the object and the insertion board, aiming to consistently evaluate the adaptability of different policies in handling various grasping and insertion points.

\paragraph{Multi-object tasks.} For the multi-object task, the assembly components for each board are placed in one of five specified starting arrangements within the designated grasp randomization area, as illustrated in Fig.~\ref{fig:board_grasp_init}. A successful policy must choose the intended piece to grasp amidst the presence of other items within the same vicinity. 
However, the board is fixed to the workspace within the insertion randomization region to reduce the complexity required during the insertion phase.

The precise protocols for each individual skill and the multi-stage tasks can be found on our website: \url{https://functional-manipulation-benchmark.github.io/procedure/index.html}.

%% file: sections/usage.tex
\section{An Imitation Learning System for the FMB} \label{sec:usage}

One significant benefit of the FMB framework is its ability to function as a standardized ``toolkit" for researchers, facilitating a convenient and unified starting point for studying various robot learning challenges. In this section, we will describe an imitation learning system we built for the FMB that serves both to provide baseline performance and a collection of components that researchers can extend to study the FMB tasks. In the next section, we analyze the performance of this system and various baselines and ablations.

\begin{figure*}[ht]
    \centering
    \begin{subfigure}[t]{0.6\linewidth}
        \includegraphics[width=\linewidth]{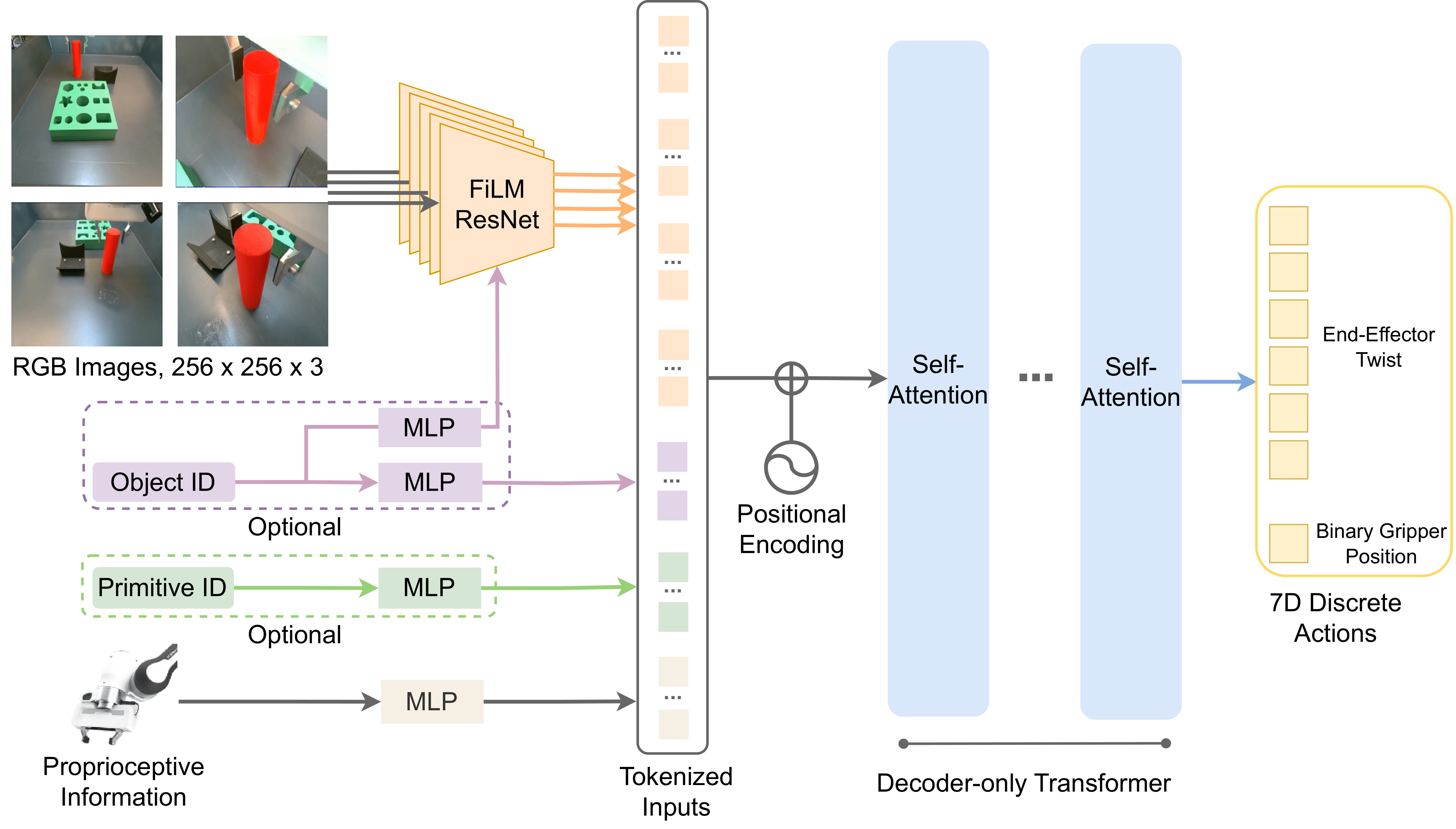}
        \caption{Architecture of the Transformer model used to train the baseline policies.}
        \label{fig:transformer}
    \end{subfigure}
    \hfill
    \rule[-.1ex]{0.7pt}{15.5em}
    \hfill
    \begin{subfigure}[t]{0.34\linewidth}
        \includegraphics[width=\linewidth]{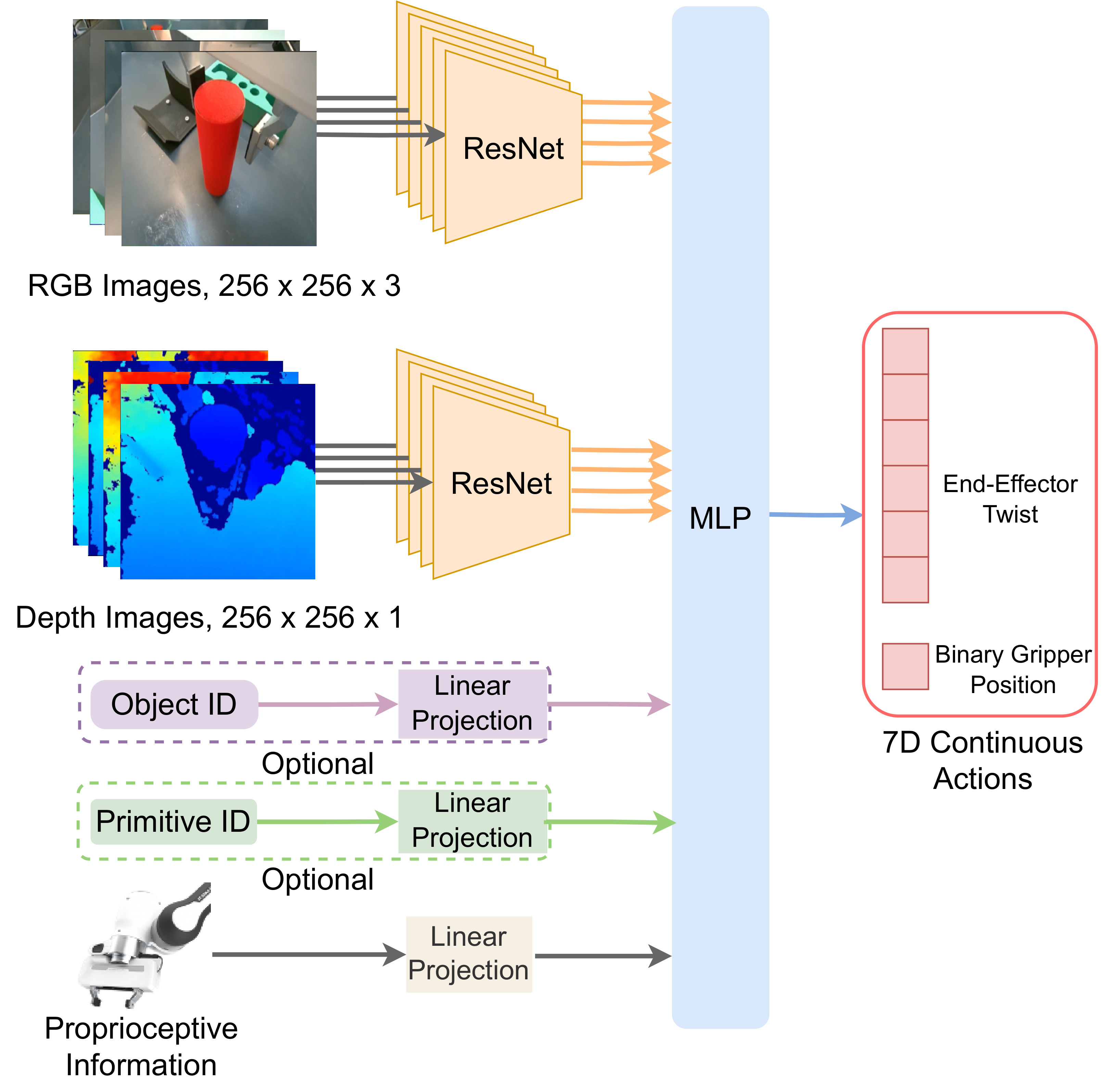}
        \caption{Architecture of the ResNet model used to train the baseline policies.}
        \label{fig:resnet}
    \end{subfigure}
    \caption{Architecture diagrams of the baseline policies. Both models encode each image view with weight-shared ResNet encoders before concatenating with proprioceptive information and optional Object and Primitive ID features to predict 7DoF actions.}
    \label{fig:networks}
    \vspace{-0.3cm}
\end{figure*}

\subsection{Imitation Learning Policies for Individual Skills}
By using the FMB dataset together with an evaluation protocol described in Sec.~\ref{sec:protocol}, we trained and tested various imitation learning models, detailed below, on individual manipulation skills. 
As we will discuss in Sec.~\ref{sec:experiments}, we also study the most effective sensor modalities for each skill as well as how the performance scales with the data available.
In all our experiments, we use two types of architectures to learn imitation learning policies, ResNet~\citep{resnet} and Transformer~\citep{vaswani2017}. In this section, we describe the detailed architectures of our imitation learning policies.

\paragraph{ResNet-based policy.}
Our ResNet-based policy's overall architecture can be seen in Fig.~\ref{fig:resnet}. 
It is composed of ResNet-34 vision backbones and an MLP as the policy head, representing a Gaussian distribution. 
We use this general structure for all of our tasks, only adapting the inputs specific to each task.
It takes multiple RGB and depth images and encodes them separately with weight-shared ResNets before concatenating the features. It also takes the robot's proprioceptive information, such as end-effector pose, twist, or force/torque measurements, and then performs linear projection before being fed into the MLP. Furthermore, the system is capable of conditioning on both the object ID and manipulation skill ID, which are represented as one-hot vectors. This mechanism is crucial for employing a hierarchical approach to effectively address long-horizon, multi-stage tasks. The output is a 6D end-effector twist as well as a binary variable that indicates whether the gripper should open or close.
In our experiments, we vary the input space to fit the needs of each scenario -- for example, when evaluating a single-task policy, the skill ID is omitted, and when evaluating the importance of force/torque measurements, we vary whether or not they are included in the input.

\paragraph{Transformer-based policy.}
Several recent works~\citep{rt12022arxiv,brohan2023rt2,rtx} showed that high-capacity models such as Transformers~\citep{NIPS2017_3f5ee243} can be effective in robotic control. The major advantages of these models lie in handling multi-modal inputs and scaling with large, diverse datasets. Our decoder-only Transformer architecture is shown in Fig.~\ref{fig:transformer}. We use weight-shared ResNet-34 encoders to tokenize images from multiple camera views. We additionally add FiLM~\citep{perez2018film} layers to condition on the object ID or primitive ID if they are required as part of the inputs to the policies. This prevents the one-hot ID vectors from being ignored by the neural network, thus making the conditioning procedure more stable. Robot proprioceptive information is tokenized via an MLP separately. These tokens, after being concatenated together with sinusoidal position embeddings, are then processed through self-attention layers with four attention heads and four MLP layers. The network outputs a discretized action consisting of a 6D end-effector twist
as well as a binary variable indicating whether the gripper should open or close. Each dimension of the continuous 6D robot action space is discretized into 256 bins during training by using a Gaussian quantizer. The discretized action space is converted back into continuous values when sending commands to the robot at runtime.

\subsection{Composing Skills to Solve Long-Horizon Tasks}
An important part of the FMB consists of the two long-horizon sequential manipulation tasks. One way of solving such tasks is to just train a ``flat'' imitation learning policy on the long-horizon trajectories.
However, this would suffer from compounding error issues~\citep{dagger}, potentially requiring a significant amount of data to achieve desirable performance.
Alternatively, we can perform the long-horizon task by employing hierarchical methods to compose individual manipulation skills with a high-level policy. In our experiments, we simply used a human-provided sequence of steps to trigger associated low-level primitives in time.
This ``human oracle'' can sequence a set of primitives to generate recovery behaviors, thus reducing compounding errors. For example, the robot can repeatedly execute the grasping primitive until the object is securely held, or opting to use a repositioning primitive to adjust the object's pose after unsuccessful grasping attempts, thus simplifying subsequent attempts. This can be achieved by using our ResNet or Transformer policy architectures with the proposed conditioning mechanism.  
Future work could explore learning such high-level policies that dynamically choose the best primitives based on the current observations. Such tasks necessitate explicit reasoning of the spatial relationships between objects and the associated manipulation skills, facilitating the use of a suitable abstract representation.

%% file: sections/experiment.tex
\section{Experiments}\label{sec:experiments}
Our experiments study the performance of the imitation learning system described previously in order to compare different variants of the imitation learning approach, understand the properties and challenges of the FMB tasks, and study the impact of different input modalities and design decisions. Specifically, our experiments study the following research questions: (1) How do various imitation learning techniques perform in our tasks so we can establish stable baselines? (2) What do the failure modes of these methods suggest about the challenges of the FMB tasks?  (3) How does the difficulty of the various FMB tasks change with the choice of input modality and policy architecture? (4) How do hierarchical policies compare to ``flat'' policies on long-horizon tasks?

To achieve this, we train a set of imitation learning policies with either ResNet~\citep{resnet} or Transformer~\citep{vaswani2017} architectures, shown in Fig. \ref{fig:networks}. We also combine these architectures with techniques such as diffusion~\citep{chi2023diffusionpolicy} and action chunking~\citep{zhao2023learning}. We'll detail these choices in the section. All pre-trained model checkpoints associated with experiments in this section can be found on our website.

\subsection{Grasping Task}

\begin{figure}[h]
    \centering
    \includegraphics[width=\linewidth]{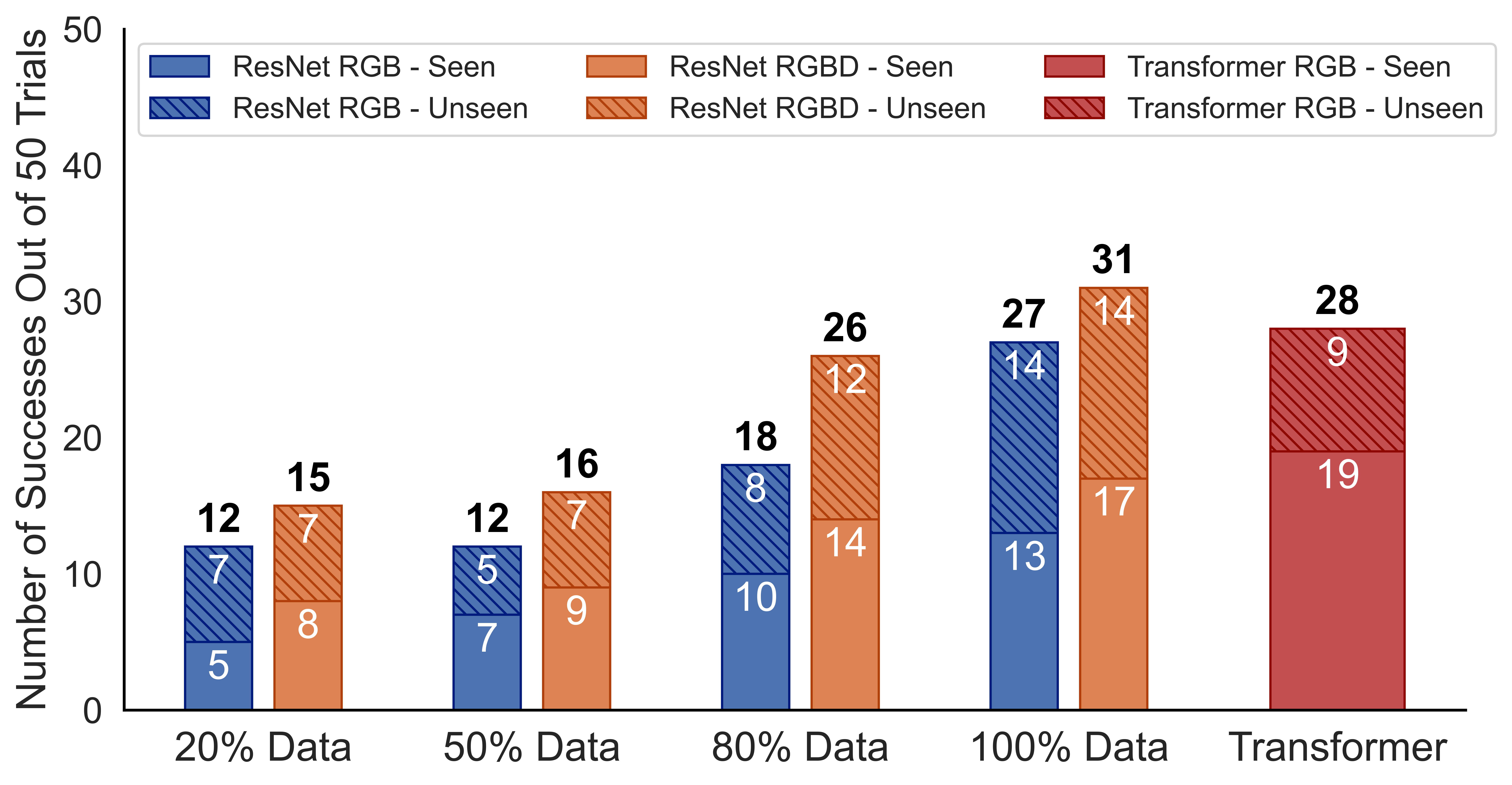}
    \caption{Number of successful grasps out of 50 trials across five seen and five unseen objects for ResNet and Transformer policies trained on various observation spaces and data percentages. The policies are able to grasp unseen objects with similar success rates as seen objects, while the overall success rate grows with the amount of training data. Training ResNet policies with depth information increases the performance across the board. 
  }
    \label{fig:grasp}
\end{figure}

An important aspect of FMB is to study the generalization across objects' physical attributes and their locations.
We conduct the grasping task to get baseline numbers of our imitation learning system, as well as to verify that we can study the proposed generalization. To achieve this, we prepare different training datasets by randomly extracting portions of data from the diverse pool of grasping data available. Specifically, we sample 20\%, 50\%, 80\%, and 100\% of the overall grasping data and study the policy's performance with the randomized evaluation procedure mentioned in Section~\ref{sec:protocol}. 
To test the generalization across objects, we conduct evaluations for both objects in the FMB dataset as well as unseen objects illustrated in Fig.~\ref{fig:novel} in accordance with our evaluation protocol detailed in Sec.~\ref{sec:protocol}.

For this task, we train both the ResNet and Transformer-based policies on RGB inputs to assess the general completion rate of the task. The specific input modality includes RGB images and TCP velocity.
To test if depth information is helpful for the grasping task, we additionally train the ResNet policy with depth alongside RGB. 
We test each policy by evaluating it on five objects in the training set and five unseen objects shown in Fig.~\ref{fig:novel}, for 5 trials each, and report the performance over the 50 trials.

Summarized in Fig. \ref{fig:grasp}, the ResNet policy's performance generally scales with the amount of training data. The Transformer policy trained on all grasping data with RGB inputs is able to grasp the objects 28 out of the 50 times tested. The ResNet policy trained on the same data and observation achieves a comparable 27 out of 50 success. The policy performance drops to 12 out of 50 as the amount of training data decreases to 20\%. 
It is interesting to note that the ResNet policies are able to generalize and grasp unseen objects shown in Fig.~\ref{fig:novel} with comparable success as seen objects regardless of the amount of training data. Furthermore, we find that depth information is beneficial as the ResNet policies trained with both depth and RGB information consistently outperform RGB-only policies trained with the same number of data. The common failure modes of this task include missing the objects and not closing the gripper at the right time.

\subsection{Repositioning Task}
\begin{figure}[h]
    \centering
    \includegraphics[width=\linewidth]{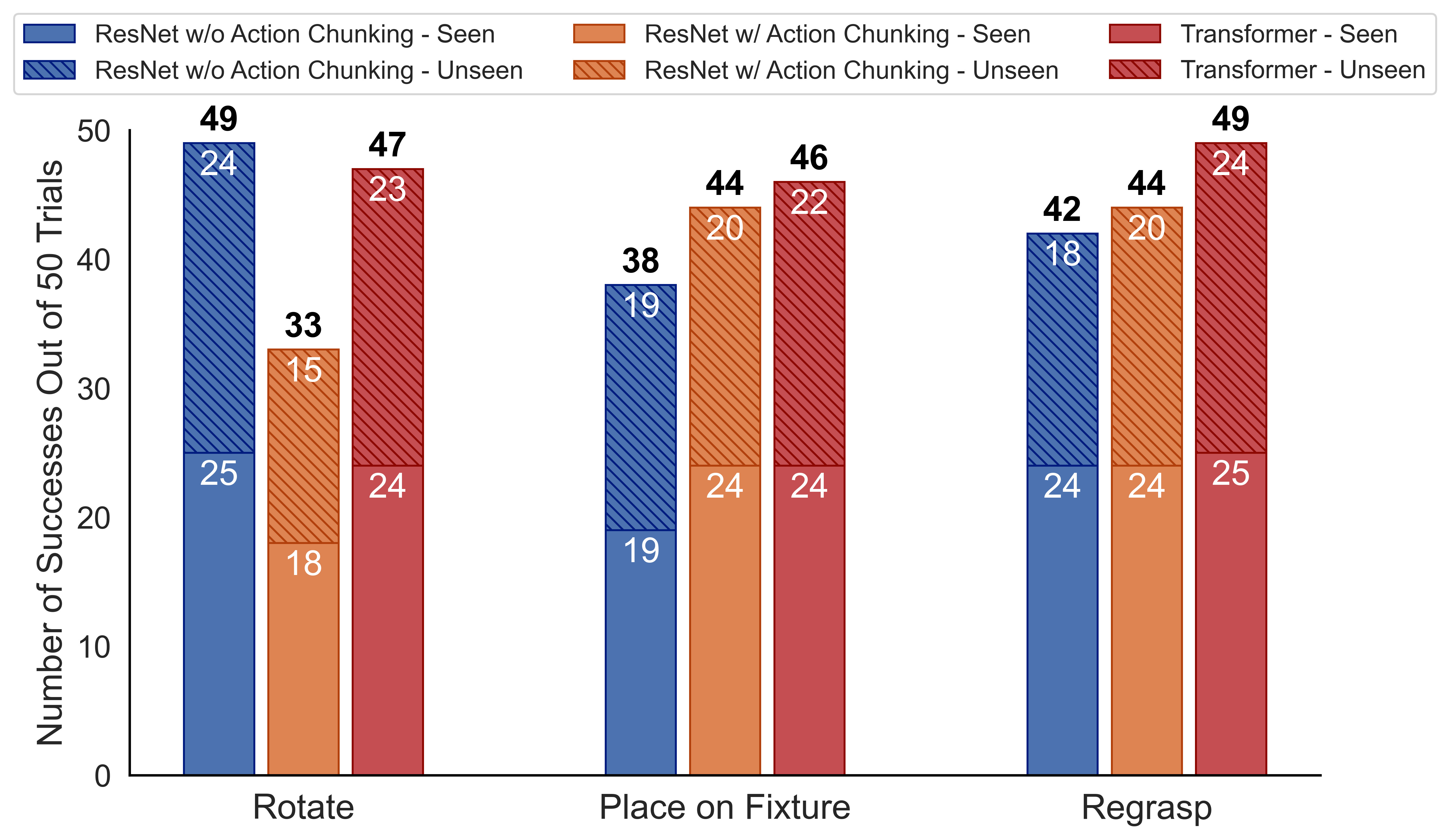}
    \caption{Number of successes for policies trained on the three repositioning tasks: Rotate, Place On Fixture, and Regrasp. We tested three models: ResNet without action chunking, ResNet with action chunking length 3, and transformer without action chunking. Each policy is evaluated 50 times across 5 seen and 5 unseen objects.}
    \label{fig:reorient}
\end{figure}

For the three repositioning skills, \texttt{rotate}, \texttt{place on fixture}, and \texttt{regrasp}, the human demonstration data can induce multiple modes of actions given the same observation. For example, an object can be rotated towards left or right contingent upon the context derived from its observational history.
We thus also train the ResNet-based policies with action chunking~\citep{zhao2023learning}, a recent method of showing promising performance handling multi-modalities in human demonstrations. 
We tested the performance of ResNet policies with and without action chunking, along with a Transformer-based policy without action chunking on seen and unseen objects, the results are presented in Fig.~\ref{fig:reorient}. The ResNet policy without action chunking outperforms its counterpart with action chunking and Transformer on the rotate skill. In contrast, the Transformer policies outperform ResNet policies with or without action chunking for the place on fixture and regrasp skills. The common failure modes for this task include not opening or closing the gripper at the right time and rotating in the wrong direction.

\subsection{Insertion Task}

\begin{table}[h!]
\centering
\begin{tabular}{p{5.4cm} c}
\toprule
\textbf{Observation} & \textbf{Success Rate} \\
\small{Three RGB,  Pose,  Vel}  & 2/25 \\
\small{Three RGBD, Pose,  Vel} & 2/25 \\
\small{Three RGB, Pose, Vel, Force/Torque} & 11/25 \\
\small{Three RGBD, Pose, Vel, Force/Torque} & 5/25 \\
\bottomrule
\end{tabular}
\caption{Ablation on input modality for insertion policy. For all policies, we include one RGB side view, two RGB wrist views, and velocity. We experimented with adding depth information from each view and adding force/torque information. By evaluating 25 trials across 5 different object sizes, we can conclude that force/torque information is crucial for contact-rich manipulation tasks like this and that depth information deteriorates performance. }
\label{tab:insert-modality}
\end{table}

For the insertion task, we studied the effect of different observation spaces of different input modalities, experimented with training a single policy for all insertion object shapes, and compared the performance of policies only trained on particular shapes.

We first experimented with the policy's input modality by training ResNet policies on rectangular object insertion data, ablating the use of depth maps as well as force/torque information. As presented in Table \ref{tab:insert-modality}, we found that the input modality has a large impact on the insertion performance, with the best being two wrist camera RGB views, one side camera RGB view, TCP pose, velocity, and force/torque. This shows that force/torque information is crucial for these contact-rich tasks, as the policy is able to tell whether the objects are in contact and execute a searching behavior. Surprisingly, using depth deteriorated the insertion performance. This could be because TCP pose information is already present in the observation space, and the noisy depth information does not aid in this precise task. Instead, it confuses the accurate end effector pose readings.
For the following experiments, we will utilize the input modalities that have led to the best performance, as demonstrated in this table.

\begin{figure}[t]
    \centering
    \includegraphics[width=\linewidth]{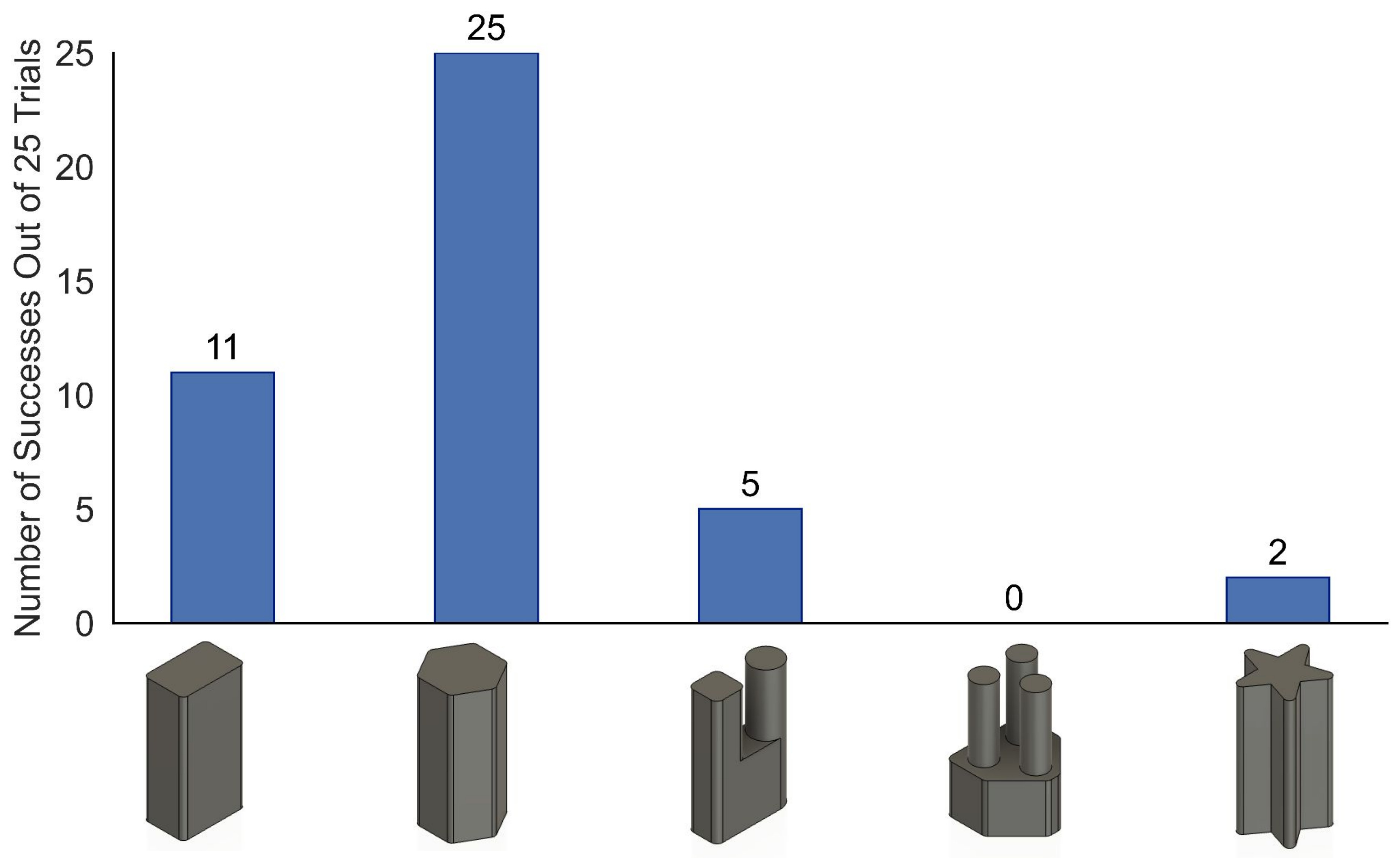}
     \caption{\textbf{Left to right}: rectangle, hexagon, circle-square, three-prong, star. We show the number of successful insertions out of 25 trials across 5 different object sizes for ResNet policies trained on individual shapes. Notice that the success rate can vary dramatically depending on the geometry of the object, creating a gradient of difficulties, which is ideal for a benchmark. }
    \label{fig:insert_shape}
\end{figure}

To carry out an initial study to understand the complexity of the insertion task, we train different ResNet policies for each object shape and evaluate them according to the procedure in 
Section~\ref{sec:protocol}. We can see that the success rate does decrease as the shape becomes more complex, with the hardest one being the three-prong object shown in Fig. \ref{fig:insert_shape}. The common failure modes include getting stuck near the holes, impeding fine-grained adjustments, difficulty in locating the matching openings, and challenges in handling multi-modalities in the demonstration data. For example, the two-pronged object with asymmetrical shapes may require a rotation between 0 to 90 degrees, depending on its grasping pose, to align with the shapes of the hole openings. 
This implies the assembly task is indeed a challenging robotic manipulation task for future benchmarking.

\begin{table}[t]
\centering
\begin{tabular}{p{5.4cm} c}
\toprule
 \textbf{Policy} & \textbf{Success Rate}\\
\small{Unconditioned ResNet} & 7/45 \\
\small{Object ID Conditioned ResNet} & 14/45 \\
\small{Object ID Conditioned Transformer} & 27/45 \\
\bottomrule
\end{tabular}
\caption{We train policies on all insertion data and evaluate 5 trials for each of the 9 object shapes. We find that using one-hot vector embedding according to the shape of the object being assembled helps the policy spatially separate the target insertion position.}
\label{tab:insert-all}
\end{table}

To study if co-training with data from other shapes helps, we then perform experiments on training policies with the insertion data that contains all the shapes and sizes.
Table \ref{tab:insert-all} shows that, when we naïvely train an unconditional ResNet policy with all the data, the policy achieves a success rate of only 7 out of 45 across 9 shapes. 
The main failure mode is trying to insert the objects into the wrong slots as well as struggling with the fine-grained execution of the insertion when in close proximity to the slots. This is not unreasonable because the policy needs to infer the right matching opening from the camera inputs, together with predicting the fine motor commands to perform the precise insertion.
The combined complexities of these tasks significantly heighten the challenge beyond that of any individual component. When we provide the policy with a one-hot vector indicating the object shape, the performance increases to 14 out of 45. Qualitatively, the policy sometimes goes to the wrong opening and sometimes fails to insert the object after going to the vicinity of the correct hole. When we train a Transformer policy with the same object shape conditioning, it achieves a 27 out of 45 success rate. We hypothesize that with the attention mechanism and the FiLM conditioning layer, our transformer policy architecture is able to pay more attention to the shape conditioning and, therefore, never reaches for the wrong hole. 

\begin{figure*}
    \centering
    \includegraphics[width=1\linewidth]{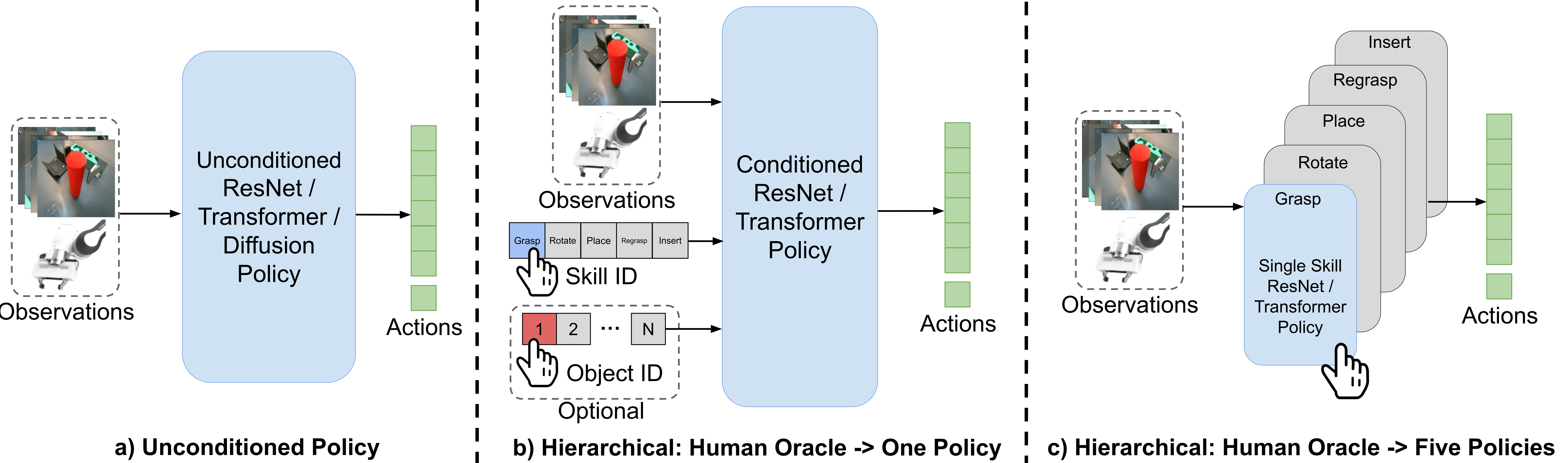}
    \caption{Illustration of the policies tested on the Multi-Stage Task. \textbf{a)} an unconditioned policy is trained on the end-to-end task. \textbf{b)} a task-conditioned policy is trained on multiple skills, and a human oracle provides the appropriate skill ID, and optionally object ID, sequentially. \textbf{c)} 5 unconditioned policies are trained on the 5 skills separately, and the human oracle selects the best policy to execute sequentially}
    \label{fig:hierarchical}
\end{figure*}

\begin{table*}[h!]
\centering
\begin{tabular}{ l C{1.8cm} C{1.8cm} C{1.8cm} C{1.8cm} }
\toprule
\textbf{Method} & \textbf{Hexagon} \newline (10 Trials) & \textbf{Circle - Square} \newline (10 Trials) & \textbf{Three - Prong} \newline (10 Trials) & \textbf{Total} \newline (30 Trials) \\

\textit{Diffusion ResNet} \\
\vspace{0.1cm}
\hspace{0.3cm}Unconditioned Policy & 0 & 0 & 0 &  0 \\

\textit{ResNet} \\
\hspace{0.3cm}Unconditioned Policy& 0 & 0 & 0  &  0 \\
\hspace{0.3cm}Hierarchical Policy\\
\hspace{0.6cm}Human Oracle $\rightarrow$ One Policy (Skill ID conditioned) & 0 & 0 & 0  &  0 \\
\vspace{0.1cm}
\hspace{0.6cm}Human Oracle $\rightarrow$ Five Policies (One policy per skill)  & \textbf{9} & \textbf{8} & 1  & 18  \\
\textit{Transformer} \\
\hspace{0.3cm}Unconditioned Policy & 0 & 0 & 0  &  0 \\
\hspace{0.3cm}Hierarchical Policy\\
\hspace{0.6cm}Human Oracle $\rightarrow$ One Policy (Skill ID conditioned) & 7 & 6 & \textbf{2}  &  15 \\
\hspace{0.6cm}Human Oracle $\rightarrow$ Five Policies (One policy per skill) & \textbf{9} & \textbf{8} & \textbf{2}  & \textbf{19}  \\
\bottomrule
\end{tabular}
\caption{We conducted an evaluation of various policies for Single-Object Multi-Stage Manipulation Tasks, focusing on the performance of Transformer and ResNet models across three distinct shapes. Notably, all unconditional policies, including those trained with diffusion models, recorded a zero success rate. We compared two types of hierarchical policies differentiated by the conditioning mechanism between the high-level and low-level policies. We found the Transformer-based policy achieved the most compelling results while providing a flexible structure for handling different input modalities.}
\label{tab:longhorizonpeg}
\end{table*}

\begin{table}[h!]
\centering
\begin{tabular}{p{5.6cm} C{2cm}}
\toprule
\textbf{Method} & \textbf{Assembly Board One} \newline (10 Trials)\\

\textit{Diffusion ResNet} \\
\hspace{0.3cm}Unconditioned Policy & 0 \\

\textit{ResNet} \\
\hspace{0.3cm}Unconditioned Policy& 0 \\
\hspace{0.3cm}Hierarchical Policy & 5\\
\textit{Transformer} \\
\hspace{0.3cm}Unconditioned Policy & 0  \\
\hspace{0.3cm}Hierarchical Policy & \textbf{7} \\
\bottomrule
\end{tabular}
\caption{We conducted an evaluation of various policies for Multi-Object Multi-Stage Manipulation Tasks, focusing on the red board as shown in Fig.~\ref{fig:teaser_fig}. The hierarchical policies use a human oracle as the high-level policy, sequentially triggering a low-level policy with the appropriate primitive and object IDs for each stage. Similar to single-object manipulation tasks, all unconditioned policies achieved zero success. Remarkably, the Transformer-based policy outperformed others, achieving a success rate of 7/10.}
\label{tab:long_horizon_board}
\end{table}

\subsection{Multi-Stage Manipulation Tasks}
As described in Sec.~\ref{sec:fmb}, the difficulties of the multi-stage assembly tasks mainly come from dealing with compounding errors introduced by each stage of manipulation, as well as reasoning the manipulation sequences.
To verify these points so as to facilitate the use of proposed hierarchical policy structures, we train ``flat" end-to-end imitation learning policies directly on the full long-horizon demonstrations. We train both ResNet and Transformer policies on all the RGB camera views together with other necessary robot proprioceptive information.  The goal of trained policies is to successfully grasp, reorient, and perform assembly. 
We assess the performance of the trained policies by conducting 10 trials for each object shape in the case of the single-object task and 10 trials for each initial object configuration in scenarios involving multi-object manipulation.

Table.~\ref{tab:longhorizonpeg} and Table.~\ref{tab:long_horizon_board} present results for both single-object and multi-object tasks, as illustrated in Fig.~\ref{fig:single-object-filmstrip} and Fig.~\ref{fig:multi-object-filmstrip}. 
In the single-object manipulation task, both the ResNet and Transformer models recorded a success rate of 0/10 when evaluated on objects of three distinct shapes. Similarly, in the multi-object manipulation task, both of them achieved a success rate of 0/10. 
The observed failure modes encompassed errors such as positioning the objects incorrectly, executing inappropriate gripper actions, and generating entirely irrational robot movements. While these outcomes serve as plausible indicators of the previously mentioned issue of error compounding, they do not entirely rule out a significant confounding factor, namely, the multi-modalities present in the human demonstration data. To further investigate this, we also train a diffusion policy~\citep{chi2023diffusionpolicy} using a ResNet. This approach models the conditional action distribution with diffusion models, which already shows promising results in representing complex multi-modal distributions of human demonstration data. However, as the results presented in Table.~\ref{tab:longhorizonpeg} and Table.~\ref{tab:long_horizon_board}, diffusion policies achieved 0/10 on both tasks. These experimental results confirmed FMB multi-stage manipulation tasks are indeed challenging, and error-compounding issues must be addressed appropriately to fully solve these tasks; which necessitate the use of hierarchical policies. 

We studied two ways of instantiating such hierarchical methods as presented in Fig.~\ref{fig:hierarchical}.
In both cases, we employ a high-level human oracle that functions as a state machine, determining the appropriate low-level skill to execute.
This oracle maintains a sequence of skills to be executed at each decision point. It is also responsible for re-executing any primitive skill that either failed in the previous step or the resulting state is deemed unsuitable for the subsequent step. For example, it may retry grasping if the object was initially grasped at a location unfavorable for insertion. The procedure is designed to terminate under two conditions: either when an unrecoverable state is encountered or when a pre-set maximum number of trial steps is reached.
While they use the same high-level policy, these approaches diverge in their representation of low-level skills. 
To assess the efficacy of the conditioning mechanism integrated into the architecture depicted in Fig.~\ref{fig:networks}, we conducted a comparative study. This involved training five distinct policies, each representing a specific low-level skill, which were then directly invoked by the high-level policy. 

First, we observed that the hierarchical policies attained measurable levels of success, in contrast to the flat policies, which demonstrated zero success as in 
Table~\ref{tab:longhorizonpeg} and 
Table~\ref{tab:long_horizon_board}. 
However, despite employing a human oracle as the high-level policy endowed with a profound understanding of the tasks to make near-optimal decisions, the maximum success rate achieved was only 19 out of 30 for single-object tasks and 7/10 for multi-object tasks. 
This indicates the inherently complex challenges presented by the FMB, affirming its suitability as a benchmark for developing advanced robotic learning methods.

For the single-object task as presented in Table ~\ref{tab:longhorizonpeg}, the Transformer-based policies achieve comparable performance between the two aforementioned hierarchical methods, namely, 19/30 compared to 15/30. However, for the ResNet-based policies, conditioned ResNet achieved zero success out of 30 trials, whereas chaining separate policies attained an 18/30 success rate, which is comparable to that of the Transformer-based policies. For the multi-object task presented in Table \ref{tab:long_horizon_board}, the conditioned hierarchical ResNet policy achieved 5/10 success compared to the conditioned hierarchical Transformer policy's 7/10 success rate. To understand this phenomenon, we found that the primary factor that causes performance difference is the ability to handle multi-modal sensory inputs between ResNet and Transformer policies. 
For each skill, there is an optimal set of sensory inputs. For example, the insertion skill reached its peak performance using three RGB camera views, supplemented with additional sensory data, as outlined in Table~\ref{tab:insert-modality}. However, we observed that incorporating a fourth camera view, specifically the right-side camera, into a ResNet policy significantly impairs its performance. This decline is primarily due to the randomized positions of the assembly board. The distant camera struggles to precisely locate the corresponding holes, leading to incorrect spatial feature associations, such as the board's edge, rather than the target location. This observation is further corroborated by the fact that incorporating a fourth camera view in multi-stage tasks, as detailed in Table.~\ref{tab:long_horizon_board}, did not adversely affect performance. This is largely attributable to the fixed position of the assembly board. In such scenarios, the redundant information provided by the additional camera remains consistent, making it sufficiently apparent for the system to effectively ignore it.
Similarly, the grasping skill generally does not benefit from adding end-effector force/torque information as it does not perform contact-rich fine-grained manipulation. 
In fact, we selected distinct sets of sensory inputs to tailor the specific requirements of each task and supplied these to five different ResNet policies. On the other hand, we fed all available sensory inputs to the conditioned policies. These policies are then required to learn the skill of selecting the appropriate set of input modalities, guided by supervision from their respective actions.
The performance of the ResNet-based policies was observed to degrade due to their difficulty in disregarding task-irrelevant inputs, leading to incorrect feature associations.
In contrast to the ResNet-based policies, the Transformer-based policies learned to effectively ignore task-irrelevant modalities, such as the non-essential fourth camera in the insertion task.
This attribute is particularly beneficial in the multi-stage, multi-task imitation learning settings characteristic of FMB tasks.

%% file: sections/discussion.tex
\section{Discussion and Limitations}\label{sec:discussion}
In this paper, we present the Functional Manipulation Benchmark (FMB).
Through the careful design of tasks, the provision of a comprehensive dataset and reproducible hardware and software system, FMB enables studying several critical challenges in robotic manipulation learning: complexity of task and skills, generalization across varied objects, and reproducibility of research.

One of the primary contributions of FMB is its focus on the complexity of manipulation tasks and the need for generalization. The tasks, ranging from single-object manipulation to complex multi-object multi-stage assemblies, capture important aspects of real-world manipulation challenges.

The inclusion of diverse 3D-printed objects enhances the need for robots to generalize their learned skills to new and unseen objects, as well as easing the burden of reproducing the proposed tasks. Our open-sourced imitation learning system, complemented by a comprehensive analysis of our experimental findings on FMB tasks, offers a foundation for researchers seeking to develop and enhance their methodologies.

Researchers can get started with FMB by first replicating our publicly available setup and trying out some of our pre-trained models. We anticipate that this initial exploration will pave the way for them to develop and evaluate new methods. 
For this reason, we look forward to their contributions and insights on the tasks proposed by FMB. 
Additionally, the nature of the FMB tasks is inherently conducive to ongoing development. Researchers have the opportunity to create novel 3D-printed objects and collect demonstrations, thereby enriching the FMB project. Notably, since the objects in multi-stage assembly tasks are constructed using a specific ``grammar", there is potential to incorporate a far greater variety of assembly boards than those currently present in FMB tasks.

Our hope is that FMB can serve as a user-friendly toolkit for individuals eager to delve into robot learning. Its inherent task complexity will foster the advancement of cutting-edge robot learning methodologies. We wish that the value FMB adds to the robot learning community will ultimately encourage community contributions, further supporting its ongoing development.